\documentclass[conference]{IEEEtran}
\IEEEoverridecommandlockouts
\usepackage{cite}
\usepackage{amsmath,amssymb,amsfonts}
\usepackage{graphicx}
\usepackage{amsmath}
\usepackage{booktabs}
\usepackage{url}
\usepackage{hyperref}
\usepackage{enumitem}
\usepackage{float}
\usepackage{graphicx}
\usepackage{tabularx}
\usepackage{arydshln} 
\usepackage{caption}
\usepackage{subcaption}
\usepackage{float} 

\usepackage{tikz}
\usepackage{caption}
\usetikzlibrary{shapes.geometric, arrows.meta, positioning}

\tikzstyle{block} = [rectangle, draw, rounded corners, minimum height=2em, minimum width=5em, align=center]
\tikzstyle{arrow} = [thick, ->, >=stealth]

\usepackage{algorithmic}
\usepackage{graphicx}
\usepackage{textcomp}
\usepackage{xcolor}
\usepackage[
    a4paper,
    top=19.05mm,   
    bottom=3cm,
    left=19.05mm,
    right=19.05mm
]{geometry}
\def\BibTeX{{\rm B\kern-.05em{\sc i\kern-.025em b}\kern-.08em
    T\kern-.1667em\lower.7ex\hbox{E}\kern-.125emX}}
\begin{document}

\title{REFLEX: Reference-Free Evaluation of Log Summarization via Large Language Model Judgment
}

\author{\IEEEauthorblockN{1\textsuperscript{st} Priyanka Mudgal}
\IEEEauthorblockA{
\textit{Portland State University}\\
Portland, USA \\
pmudgal@pdx.edu}
}

\maketitle

\begin{abstract}

Evaluating log summarization systems is challenging due to the lack of high-quality reference summaries and the limitations of existing metrics like ROUGE and BLEU, which depend on surface-level lexical overlap. We introduce REFLEX, a reference-free evaluation metric for log summarization based on large language model (LLM) judgment. REFLEX uses LLMs as zero-shot evaluators to assess summary quality along dimensions such as relevance, informativeness, and coherence, without requiring gold-standard references or human annotations. We show that REFLEX produces stable, interpretable, and fine-grained evaluations across multiple log summarization dataset, and more effectively distinguishes model outputs than traditional metrics. REFLEX provides a scalable alternative for evaluating log summaries in real-world settings where reference data is scarce or unavailable. Code is available at: \href{https://github.com/prmudgal/Reflex}{\texttt{https://github.com/prmudgal/Reflex}}

\end{abstract}


\section{\textbf{Introduction}}

Logs are foundational to the operation and maintenance of modern software systems. They record fine-grained information about system behavior, including internal state transitions, API calls, error traces, performance metrics, and user interactions. These logs serve as critical artifacts for developers, operators, and security teams enabling them to trace faults, detect anomalies, audit behavior, and understand system performance ~\cite{10.1145/3650212.3652123}. As distributed systems and microservices architectures grow in complexity, the scale and diversity of log data have increased dramatically, often reaching millions of entries per day in large-scale deployments.

Despite their importance, raw logs are notoriously difficult for humans to interpret. They are often verbose, unstructured or semi-structured, and filled with low-level system details or cryptic error codes. Moreover, logs typically lack high-level summaries or explanations, requiring engineers to sift through large volumes of data to locate relevant events ~\cite{10301257}. This process is time-consuming, error-prone, and mentally taxing specifically during high-pressure scenarios like production outages or security incidents. Consequently, there is a growing need for tools that can condense logs into concise, human-readable summaries that highlight the most relevant and actionable information.

Automatic log summarization aims to address this challenge by generating natural language descriptions or condensed views of raw log sequences. Traditional approaches have relied on rule-based techniques such as regular expressions, templates, or heuristics, which are often tailored to specific log formats or domains \cite{Katukam2025}. While these techniques are effective in narrow settings, they are brittle and difficult to scale. Recently, LLMs such as GPT-4 \cite{brown2020language} and BART \cite{lewis2019bart} have emerged as compelling alternatives, due to their ability to process unstructured input, infer context, and generate fluent, coherent summaries without manual feature engineering or domain-specific rules.

While LLM-based summarization shows promise, evaluating the quality of the generated summaries remains a key bottleneck \cite{10.1145/3731445}. Unlike traditional summarization tasks (e.g., news articles), log summaries often lack gold-standard references. Even when reference summaries are available, they may differ substantially in lexical form while expressing the same semantics which makes surface-level metrics like \textsc{ROUGE} \cite{lin2004rouge} or \textsc{BLEU} \cite{10.3115/1073083.1073135} unreliable. These metrics struggle to capture meaning preservation or relevance in highly variable outputs. Human evaluation, although more robust, is time-consuming, expensive, and difficult to standardize, especially in operational settings where rapid iteration and deployment are required.

We hypothesize that the reasoning capabilities of LLMs, particularly their ability to assess semantic coherence and contextual relevance, can support reference-free evaluation in the log summarization domain. Unlike standard summarization tasks, log summaries often vary significantly in lexical form while preserving the same underlying meaning, rendering overlap-based metrics like \textsc{ROUGE} ineffective. We show that prompting an LLM to directly judge (log input, log summary) pairs produces scores with strong correlation to human preferences across multiple log datasets. We refer to this evaluation process as \textbf{REFLEX} (\textbf{Re}ference-\textbf{F}ree \textbf{}Evaluation of \textbf{L}og Summarization via LLM \textbf{Ex}amination). Beyond raw correlation, we find that REFLEX captures quality dimensions such as relevance, informativeness, and fluency more consistently than traditional metrics.


Our contributions are as follows:

\begin{itemize}
    \item We propose \textbf{REFLEX}, a new reference-free evaluation framework for log summarization based on LLM judgment.
    \item We demonstrate that REFLEX produces more informative and discriminative evaluations than traditional metrics across multiple log summarization datasets.
    \item We release our evaluation protocol and analysis tools to support reproducible research in log summarization.
\end{itemize}

\section{\textbf{Related Work}}

\subsection{\textbf{Log Analysis and Summarization}}

Traditional log analysis tools such as ELK Stack and Splunk provide indexing and search capabilities, but lack semantic summarization. Rule-based methods have been employed to extract templates or anomalies (e.g., Drain, LogPai), but they struggle with generalization and require significant domain knowledge. Recent efforts have explored deep learning and large language model based log analysis \cite{MudgalChatGPT, zhang2023survey}, anomaly detection \cite{}, log classification \cite{RAMACHANDRAN20231722}, log parsing \cite{jiang2024lilac, zhong2024logparser, astekin2024comparative, xu2024help}, failure management \cite{zhang2024survey}, reasoning \cite{pan2025enhancing}, and event prediction \cite{10762255, fu2024logtransformer}, yet few focus on natural language summarization. LogGPT \cite{loggpt2023} is a recent approach that applies GPT-style models to analyze logs, but its focus is primarily on anomaly detection. Our work complements this by targeting summarization and introducing a unified evaluation method. 

\subsection{\textbf{Large Language Models for Summarization}}

Transformer-based models like BART \cite{lewis2019bart}, T5 \cite{raffel2020exploring}, and GPT-3/4 \cite{brown2020language} have set state-of-the-art benchmarks on summarization tasks. While most benchmarks involve natural text (news, reviews), few works study their performance on semi-structured inputs like logs. Flan-T5 \cite{chung2022scaling} introduced instruction tuning, making it more adaptable to unseen tasks. The closest to our work is \cite{bertalan2025clustering}, where Bertalan et al. proposed a supervised log summarization method that leverages clustering techniques using log embeddings, parsed variables, and line proximity, followed by topic modeling and word analysis to highlight important log lines, showing superior performance over existing approaches on various datasets. In our work, we leverage the strengths of BART \cite{lewis2019bart}, T5 \cite{raffel2020exploring}, and GPT-3/4 \cite{brown2020language} crafting log-specific prompts for log summarization. A recent study \cite{Katukam2025} introduced a scalable, AI-driven log summarization platform leveraging Google’s Gemini 1.5 Flash model to automate security insight extraction from diverse log formats, significantly improving SOC efficiency and threat visibility; however, the work did not include any quantitative scoring or evaluation metrics. 

\subsection{\textbf{Evaluation Metrics}}
Classic metrics like ROUGE \cite{lin2004rouge}, BLEU \cite{10.3115/1073083.1073135}, and METEOR \cite{10.5555/1626355.1626389} focus on n-gram overlap, which poorly captures semantic similarity. Sentence-BERT \cite{reimers2019sentence} offers an embedding-based approach that measures meaning similarity using cosine distance. Our method uses to compute alignment between LLM-generated and human-authored summaries.





\section{\textbf{Methodology}}

Our system implements a modular pipeline designed for both generating and evaluating log summaries using LLMs. The architecture is composed of two core components that work together to support flexible experimentation, reliable evaluation, and comprehensive benchmarking.
\subsection{\textbf{Preprocessing Layer}}

Before logs are passed to the LLM summarizer, they are processed through a Preprocessing Layer designed to normalize input structure, remove noise, and segment long sequences. Logs often vary in formatting and may include timestamps, identifiers, nested JSON structures, or plain text messages. This component standardizes such inputs to ensure consistent downstream behavior.

Key responsibilities of this layer include:

\begin{itemize}[noitemsep]
\item \textbf{Format Normalization:} Converts various log formats (e.g., JSON, syslog, Apache logs) into a unified textual representation.
\item \textbf{Field Extraction:} Selectively extracts relevant fields such as timestamps, log levels, error codes, and messages.
\item \textbf{Noise Filtering:} Removes redundant or non-informative entries (e.g., heartbeat messages, debug logs).
\item \textbf{Sequence Chunking:} Splits long log sequences into chunks that fit within the input token limits of the LLM. Chunks may be constructed based on fixed window sizes or semantically grouped using heuristics or change-point detection.

\end{itemize}

This preprocessing is essential for ensuring that the LLM receives well-structured and focused input, which directly impacts the quality and consistency of the generated summaries. It also decouples log ingestion concerns from the summarization logic, reinforcing the modularity of the overall architecture.
\begin{figure*}[htbp]
\vspace{-0.8cm} %
    \centering
    \includegraphics[width=0.9\textwidth, trim=0 150 100 0, clip]{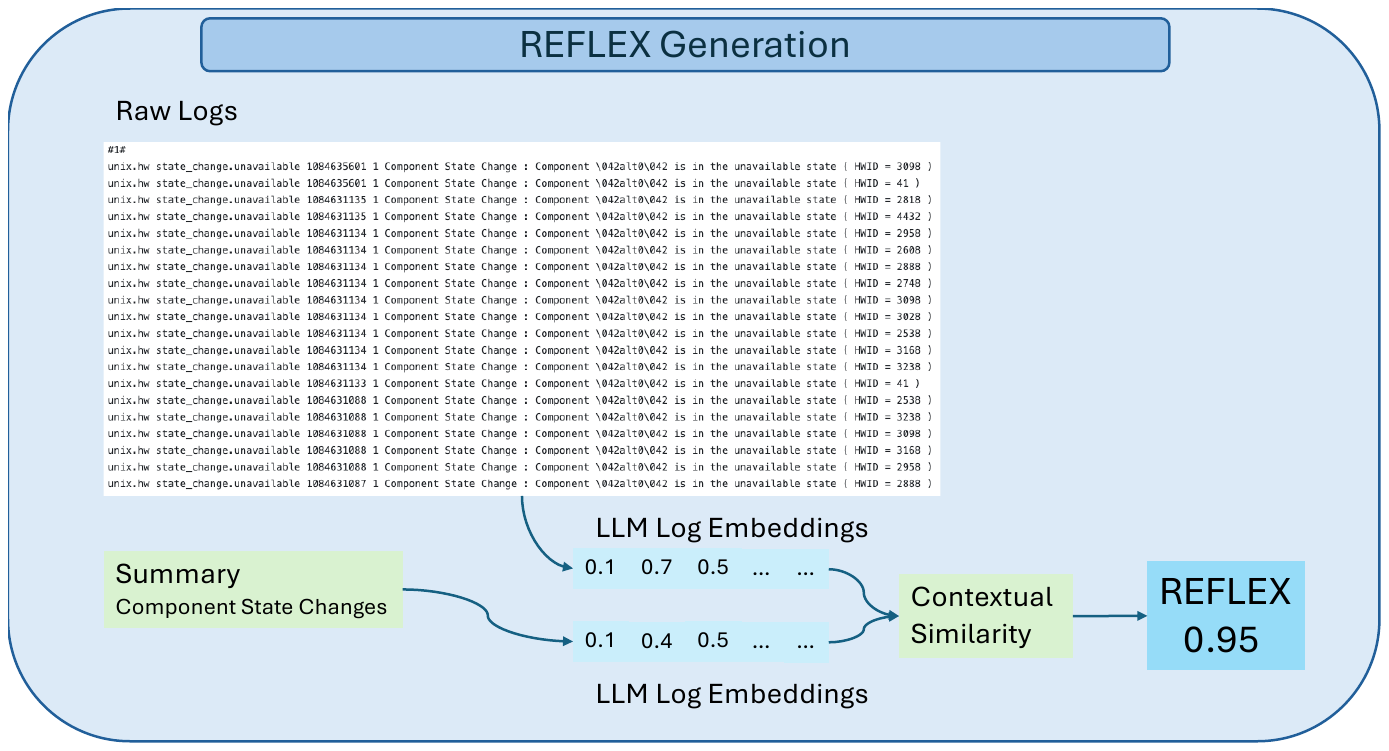}
    \caption{REFLEX uses LLM to generate summaries from logs and evaluates them automatically, without requiring human-written references, similar to how experts judge log readability.}
    \label{fig:pipeline}
\end{figure*}

\subsection{\textbf{LLM-Based Log Summarization}}

The first component of our pipeline is the \textit{LLM Summarizer}, which generates human-readable summaries from raw log sequences. These inputs may consist of unstructured or semi-structured records such as system traces, error messages, and API call logs. Given the wide variety of deployment scenarios, the summarizer is designed to be model-agnostic and easily extensible, supporting both proprietary and open-source LLMs.

We represent a log sequence as:
\begin{equation}
    \mathcal{L} = \{l_1, l_2, \ldots, l_n\},
\end{equation}
where $l_i$ denotes the $i$-th log entry in a sequence of length $n$.

The summarizer $\mathcal{S}$ is modeled as a conditional generation function:
\begin{equation}
    \hat{y} = \mathcal{S}(\mathcal{L}, p; \theta),
\end{equation}
where $\hat{y}$ is the generated summary, $p$ is the prompt, and $\theta$ are the LLM's parameters. The summary generation is conditioned on the input logs, prompt, and model parameters.

Prompts are dynamically constructed from the log sequence:
\begin{equation}
    p = \mathcal{P}(\mathcal{L}) = \text{``Summarize the following logs:\textbackslash n''} + \texttt{join}(\mathcal{L}),
\end{equation}
where the prompt includes a fixed instruction concatenated with the log content. The prompt construction is done by combining a textual instruction with the joined log lines.

To support multiple models, we define:
\begin{equation}
    \hat{y}_k = \mathcal{M}_k(\mathcal{L}, p_k; \theta_k),
\end{equation}
where $\mathcal{M}_k$ denotes the $k$-th model, $\theta_k$ its parameters, and $p_k$ its respective prompt. With this capability, it has the flexible substitution of different LLMs in the summarization process. It can interface with:

\begin{itemize}[noitemsep]
    \item \textbf{Proprietary LLMs:} OpenAI’s GPT-4 accessed through its API, which offers strong contextual understanding and fluent natural language generation.
    \item \textbf{Open-source LLMs:} BART, Flan-T5, and T5 variants accessible via Hugging Face’s Transformers library, allowing for self-hosted inference and fine-tuning.
\end{itemize}

This formalized abstraction allows for controlled comparisons between models and prompts while maintaining flexibility for real-world deployment.

\subsection{\textbf{Evaluation Engine}}

The second component, the {Evaluation Engine}, measures the quality of generated summaries by computing semantic similarity scores against human-authored reference summaries where available. This engine utilizes pretrained sentence embedding models, specifically Sentence-BERT variants \cite{reimers2019sentence}, to transform textual summaries into dense vector representations. By embedding both the LLM-generated summary and the human reference summary into a shared semantic space, the engine calculates cosine similarity scores as a proxy for semantic closeness:

\[
\text{Similarity} = \frac{E_g \cdot E_h}{\|E_g\|\|E_h\|}
\]

where \( E_g \) and \( E_h \) are the normalized embeddings of the generated and human summaries, respectively. This embedding-based evaluation abstracts away from surface-level lexical matches and focuses on the underlying semantic content, which is particularly important given the diversity of valid summary expressions in the log domain. Moreover, this design enables easy integration of alternative evaluation metrics or embedding models, supporting extensibility and experimentation with emerging semantic evaluation techniques.

\subsection{\textbf{Modularity and Extensibility}}

Our system’s modular design promotes easy swapping and customization of each component. New summarization models, embedding techniques, or evaluation strategies can be integrated with minimal changes to the pipeline. For example, integrating a new transformer model requires implementing a thin wrapper adhering to a common interface for input formatting and output parsing. Similarly, evaluation backends can be extended by registering new metric classes within the scoring engine.

This modularity supports a wide range of use cases:

\begin{itemize}[noitemsep]
\item \textbf{Research:} Enables benchmarking of novel summarization architectures or prompt tuning methods.
\item \textbf{Operations:} Allows fine-tuning or reconfiguration for domain-specific logs such as cloud infrastructure, security audits, or user telemetry.
\item \textbf{Deployment:} Facilitates adaptation to on-premise, hybrid, or fully cloud-based environments.
\end{itemize}

Through this design, the system enables reproducible experiments, comparative evaluation, and rapid iteration which are essential for both academic research and real-world applications in log analysis and summarization.

\section{\textbf{Experimental Setup}}

\subsection{\textbf{Dataset}}
To evaluate our log summarization and \textbf{REFLEX} evaluation framework, we utilized datasets from two public sources: the LogSummary dataset by Meng et al.~\cite{10017337}, and the LogHub repository ~\cite{10.1145/3650212.3652123, 10301257}. The LogSummary dataset contains raw log entries collected from real-world, production-like software systems, including HDFS, BGL, HPC, Proxifier, {ZooKeeper}, and {Spark}. Each log group in the dataset consists of approximately 20 consecutive log messages and is accompanied by both manually curated (gold) summaries and automatically generated summaries produced by the LogSummary framework. The summaries are constructed using an information extraction pipeline that includes log parsing, triplet extraction, ranking, and natural language realization.

In addition to LogSummary, we also leveraged raw log data from {LogHub} ~\cite{10.1145/3650212.3652123, 10301257}, a comprehensive repository of system log datasets collected from a variety of open-source systems, including distributed databases, cloud platforms, and big data frameworks. LogHub provides structured and cleaned log datasets that serve as a reliable foundation for both supervised and unsupervised log analysis tasks.

These datasets span multiple layers of modern software stacks—including storage, network, caching, and authentication components—and provide a diverse set of log formats and semantics. This diversity enables a robust evaluation of REFLEX's ability to assess summary quality across heterogeneous logging scenarios.

Each log entry is paired with a machine-generated summary created using LogSummary’s log parsing and ranking-based summarization pipeline. These summaries serve as reference outputs to test the effectiveness of our REFLEX framework in rating and benchmarking the quality of generated log summaries.

\begin{figure}[htbp]
\centering
\resizebox{\columnwidth}{!}{%
\begin{tikzpicture}
    \draw[thick] (0,0) rectangle (12,5);
    \node[align=left, anchor=north west] at (0.2,4.8) {\parbox{11.5cm}{\footnotesize\ttfamily
        INFO dfs.FSNamesystem : BLOCK* NameSystem.addStoredBlock : blockMap updated : 10.251.71.68 : 50010 is added to blk\_-1886295043152742159 size 67108864.\\
        INFO dfs.FSNamesystem : BLOCK* NameSystem.allocateBlock : /user/root/rand/\_temporary/\_task\_200811092030\_0001\_m\_000006\_0/\\
        part-00006. blk\_-2581653693275159104.\\
        INFO dfs.FSNamesystem : BLOCK* NameSystem.addStoredBlock : blockMap updated : 10.251.65.203 : 50010 is added to blk\_194626696959819525 size 67108864. \\
\\
        {Summary:} Receiving block src; blockMap updated; Verification succeeded.
    }};
\end{tikzpicture}
}
\caption{HDFS block update log messages and provided summary \cite{10017337}.}
\label{fig:log-multi-box}
\end{figure}

While we did not modify the original dataset, we focus our evaluation on {100 log groups}, each comprising approximately 20 contiguous log lines, as defined in the LogSummary gold standard benchmark. These cover diverse operational domains and allow us to assess how well REFLEX captures summary quality across varied log formats and content.

This dataset, although automatically summarized, represents a {realistic and heterogeneous} sample of logs encountered in distributed systems. Its structure supports robust evaluation of frameworks like REFLEX that aim to objectively measure log summary effectiveness across dimensions such as informativeness, precision, and relevance.

\subsection{\textbf{Models Evaluated}}

We benchmark three representative summarization models chosen to cover a range of model sizes, architectures, and access modalities:

\begin{itemize}[noitemsep]
    \item \textbf{GPT-4 (OpenAI API):} A state-of-the-art proprietary autoregressive LLM accessed via API, known for its strong natural language understanding and generation capabilities.
    \item \textbf{Flan-T5-xl:} An open-source, instruction-tuned encoder-decoder model with approximately 3 billion parameters, optimized for few-shot and zero-shot generalization.
    \item \textbf{BART-Large-CNN:} A transformer-based encoder-decoder model pretrained for summarization tasks, known for its efficiency and strong baseline performance.
\end{itemize}

All models are prompted or fine-tuned to generate concise summaries of log inputs, using standardized prompts where applicable to ensure fair comparison.

\begin{table*}[htbp]
\vspace{2cm}
\centering
\small
\caption{Similarity scores across different log types for three REFLEX variants and corresponding ROUGE evaluation metrics. The REFLEX models (GPT-4, BART, Flan-T5) are evaluated for their output similarity, followed by ROUGE-1, ROUGE-2, and ROUGE-L metrics to measure lexical overlap with reference summaries.}
\begin{subtable}[t]{\textwidth}
\centering
\caption{REFLEX(GPT-4) Results}
\begin{tabularx}{\textwidth}{l *{6}{>{\centering\arraybackslash}X}}
\toprule
\textbf{Metric} & \textbf{BGL} & \textbf{HDFS} & \textbf{HPC} & \textbf{Proxifier} & \textbf{Spark} & \textbf{Zookeeper} \\
\midrule
ROUGE-1           & 0.2551 & 0.1540 & 0.2647 & 0.1997 & 0.1996 & 0.2681 \\
ROUGE-2           & 0.1318 & 0.0177 & 0.0750 & 0.0317 & 0.0531 & 0.0683 \\
ROUGE-L           & 0.2329 & 0.1250 & 0.2403 & 0.1480 & 0.1596 & 0.1807 \\
\hdashline
\\[-1pt]
REFLEX(GPT-4)     & 0.5439 & 0.4681 & 0.5951 & 0.5521 & 0.4753 & 0.6354 \\
\bottomrule
\end{tabularx}

\label{tab:gpt4}
\end{subtable}

\vspace{1em}

\begin{subtable}[t]{\textwidth}
\centering
\caption{REFLEX(BART) Results}
\begin{tabularx}{\textwidth}{l *{6}{>{\centering\arraybackslash}X}}
\toprule
\textbf{Metric} & \textbf{BGL} & \textbf{HDFS} & \textbf{HPC} & \textbf{Proxifier} & \textbf{Spark} & \textbf{Zookeeper} \\
\midrule
ROUGE-1           & 0.2920 & 0.1094 & 0.1195 & 0.2029 & 0.2413 & 0.2281 \\
ROUGE-2           & 0.1993 & 0.0343 & 0.0680 & 0.1036 & 0.1166 & 0.1098 \\
ROUGE-L           & 0.2824 & 0.1016 & 0.1125 & 0.1849 & 0.2114 & 0.2013 \\
\hdashline
\\[-1pt]
REFLEX(BART)      & 0.5506 & 0.3841 & 0.3462 & 0.5024 & 0.4694 & 0.5669 \\
\bottomrule
\end{tabularx}

\label{tab:bart}
\end{subtable}

\vspace{1em}

\begin{subtable}[t]{\textwidth}
\centering
\caption{REFLEX(Flan-T5) Results}
\begin{tabularx}{\textwidth}{l *{6}{>{\centering\arraybackslash}X}}
\toprule
\textbf{Metric} & \textbf{BGL} & \textbf{HDFS} & \textbf{HPC} & \textbf{Proxifier} & \textbf{Spark} & \textbf{Zookeeper} \\
\midrule
ROUGE-1           & 0.4447 & 0.2003 & 0.4893 & 0.3496 & 0.2108 & 0.3349 \\
ROUGE-2           & 0.3276 & 0.0759 & 0.2818 & 0.1288 & 0.1000 & 0.1720 \\
ROUGE-L           & 0.4385 & 0.1938 & 0.4695 & 0.3196 & 0.2048 & 0.3204 \\
\hdashline
\\[-1pt]
REFLEX(Flan-T5)   & 0.5736 & 0.3544 & 0.7034 & 0.5619 & 0.4098 & 0.6015 \\
\bottomrule
\end{tabularx}

\label{tab:flant5}
\end{subtable}

\label{tab:all_models}
\end{table*}

\begin{figure*}[htbp]
    \centering
    \begin{subfigure}[b]{0.32\textwidth}
        \centering
        \includegraphics[width=\textwidth]{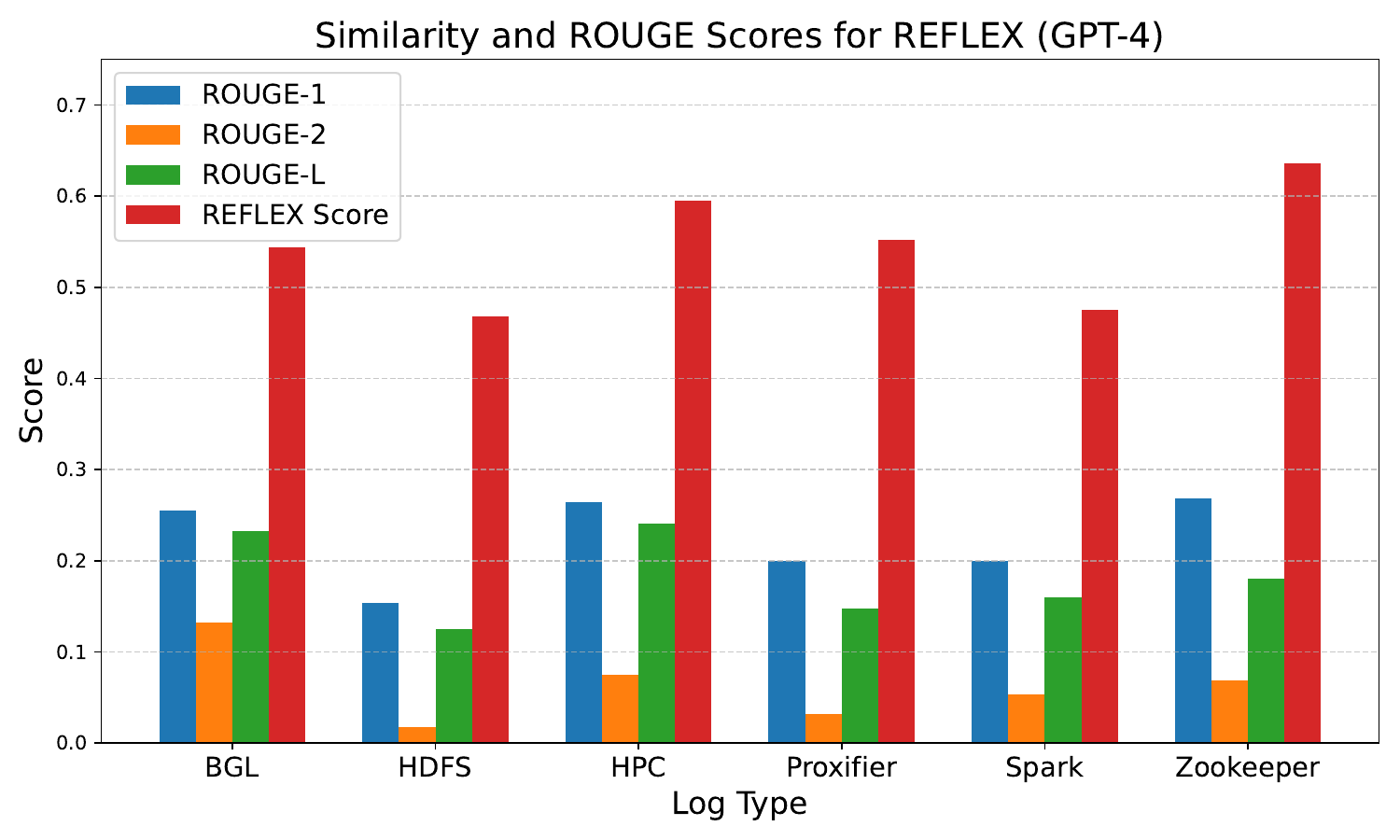}
        \caption{REFLEX (GPT-4)}
        \label{fig:reflex_gpt4}
    \end{subfigure}
    \hfill
    \begin{subfigure}[b]{0.32\textwidth}
        \centering
        \includegraphics[width=\textwidth]{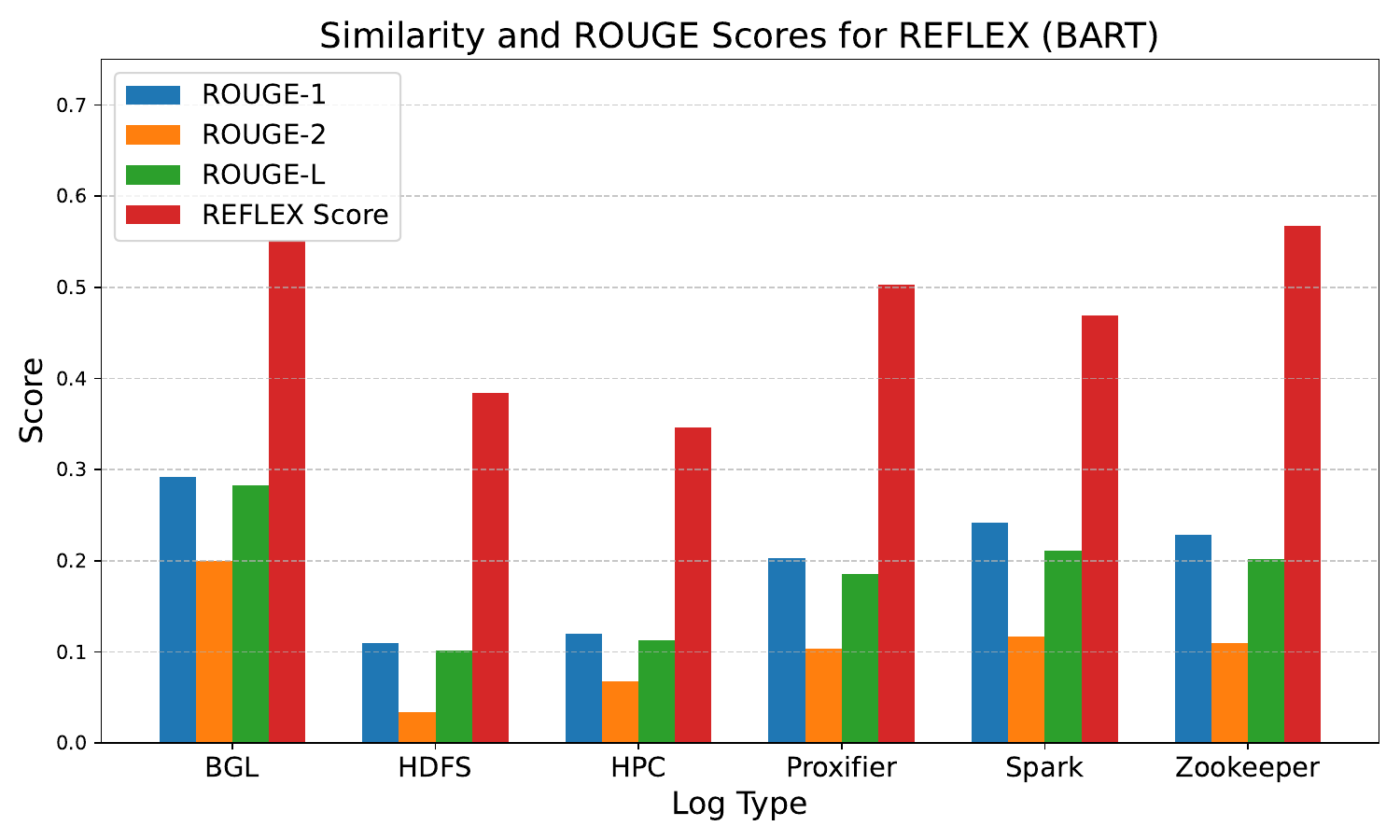}
        \caption{REFLEX (BART)}
        \label{fig:reflex_bart}
    \end{subfigure}
    \hfill
    \begin{subfigure}[b]{0.32\textwidth}
        \centering
        \includegraphics[width=\textwidth]{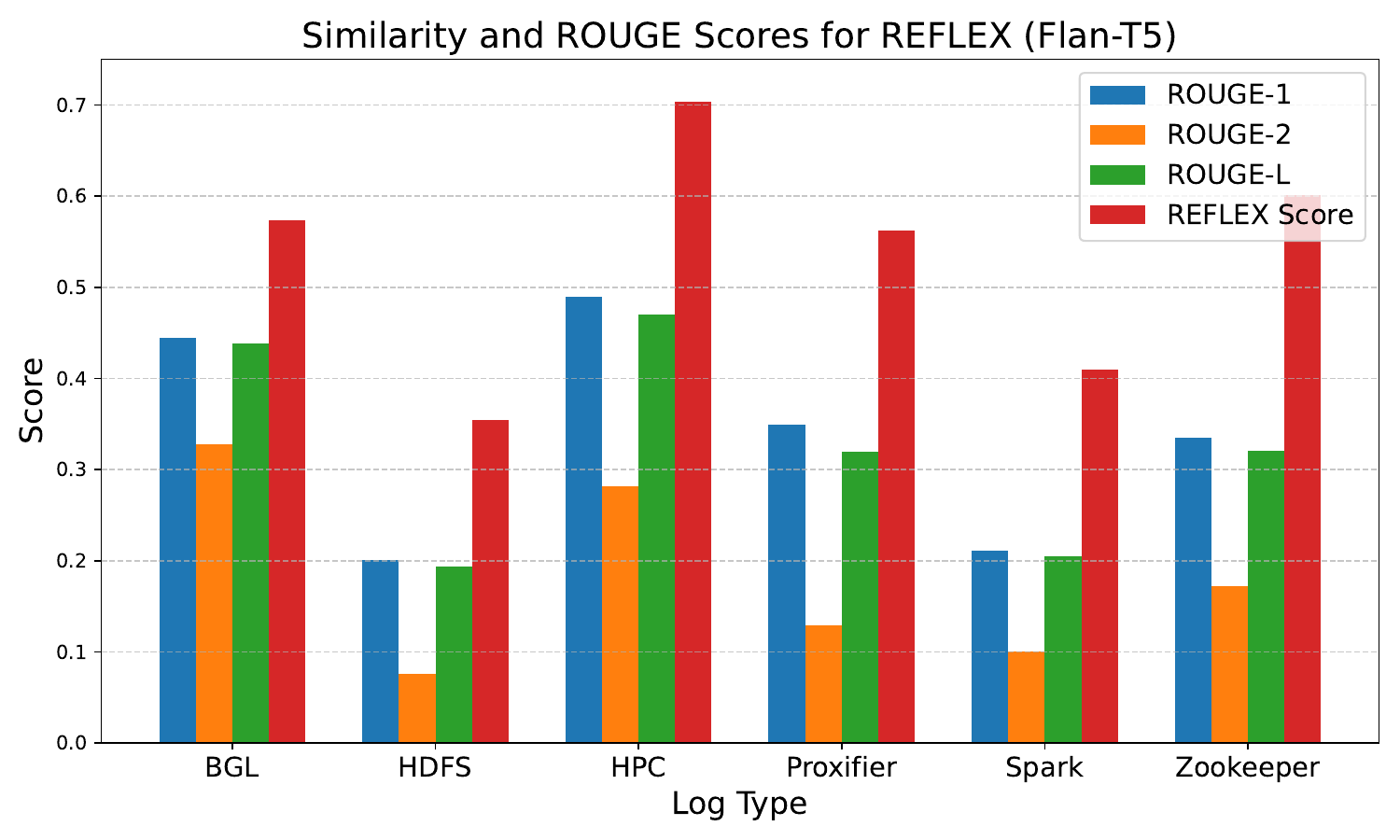}
        \caption{REFLEX (Flan-T5)}
        \label{fig:reflex_flant5}
    \end{subfigure}
    \caption{Comparison of similarity and ROUGE scores across log types for three REFLEX variants.}
    \label{fig:reflex_comparison}
\end{figure*}

\subsection{\textbf{Hardware and Runtime Environment}}

We ran experiments involving open-source models on a workstation equipped with an NVIDIA RTX 3090 GPU and 32GB of RAM, which ensured efficient execution of transformer-based models. For GPT-4 summarization, we accessed the OpenAI API and configured the settings to promote consistent output and set the temperature to 0.3 to minimize randomness. 

\subsection{\textbf{Evaluation Metrics}}

In addition to our proposed REFLEX metric, we compute and compare the following widely-used automatic evaluation metrics for summarization:

\begin{itemize}[noitemsep]
    \item \textbf{ROUGE-1:} Measures unigram (word-level) overlap between the candidate and reference text to assess recall-oriented lexical similarity.
    \item \textbf{ROUGE-2:} Measures bigram (two-word sequence) overlap between the candidate and reference text to assess recall-oriented lexical similarity.
    \item \textbf{ROUGE-L:} Measures the longest common subsequence (LCS) between the candidate and reference text to assess recall-oriented lexical similarity.

    \item \textbf{REFLEX} Measures contextual and semantic similarity using dense embeddings, serving as a baseline for REFLEX’s embedding-based evaluation.
\end{itemize}

These metrics provide a comprehensive baseline against which to validate the correlation and robustness of the REFLEX score.

\section{\textbf{Results}}

\subsection{\textbf{Quantitative Comparison}}

The results in Table \ref{tab:all_models} reveal notable variations in similarity scores across different log types and evaluation metrics. Among the REFLEX variants, Flan-T5 generally achieves the highest similarity scores, particularly on the HPC and BGL datasets, suggesting its strong performance in capturing semantic structure in those domains. GPT-4 also performs consistently well, outperforming BART across most log types. BART, while trailing the other models in REFLEX scores, shows competitive performance on Proxifier and Spark. The ROUGE scores follow a similar trend, with the highest lexical overlap observed on BGL and HPC logs, indicating that these datasets may be more amenable to structured summarization or may contain more repetitive patterns. In contrast, HDFS and Proxifier yield lower ROUGE-2 scores, reflecting their more diverse or less predictable language patterns.

\subsection{\textbf{Qualitative Examples}}
In this subsection, we present a qualitative example to illustrate the differences in summarization style and content between human-generated and model-generated summaries. The original log entry reports a database connection timeout occurring after 30 seconds, followed by an automatic retry attempt. The human summary concisely captures this event as a "Database connection timeout, retry initiated," highlighting the key points succinctly. Similarly, the GPT-4 summary effectively paraphrases the log with "Connection to the database timed out; retrying," maintaining clarity and completeness. The Flan-T5-xl model produces a shorter, more compressed summary, stating only "Timed out after 30 seconds," which omits the retry action but preserves the timeout information. Finally, the BART model closely mirrors the human summary by stating "Database connection timed out, system retrying," balancing completeness and readability. This example demonstrates how different models vary in detail and phrasing, with GPT-4 and BART providing fuller context compared to the more concise Flan-T5-xl.

\section{\textbf{Discussion}}

\subsection{\textbf{Model Behavior}}

Our experiments show that GPT-4 consistently exhibits superior contextual understanding when summarizing or rephrasing log data. It is able to capture nuanced relationships between events, timestamps, and system components, producing outputs that resemble human-written summaries. In comparison, sequence-to-sequence models such as BART and Flan-T5 demonstrate reasonable performance on structured logs but occasionally misinterpret subtle technical details. Short or fragmented log entries tend to confuse these models, leading to partially inaccurate summaries or omission of critical events. These observations suggest that while modern transformer-based models are capable of sophisticated reasoning, the depth of understanding varies with model scale, pretraining corpus, and prompt design.

\subsection{\textbf{Cost and Deployment Trade-offs}}

Deploying large language models involves a clear trade-off between performance, cost, and privacy. OpenAI models like GPT-4 offer state-of-the-art fluency and accuracy but operate through API calls, incurring token-based usage fees that can become substantial in large-scale log analysis scenarios. Moreover, sending sensitive log data to an external API introduces privacy and compliance concerns, particularly in enterprise or regulated environments. In contrast, open-source models available through Hugging Face, such as BART or Flan-T5, can be deployed locally, providing full data control and eliminating external transmission risks. While these models generally exhibit lower fluency and may require fine-tuning to reach parity with GPT-4, local deployment can be preferable in security-sensitive or cost-constrained settings. Additionally, local deployment allows for integration into existing pipelines with minimal network overhead, which can be critical for real-time log monitoring systems.

\subsection{\textbf{Limitations}}

Despite these insights, there are several limitations that should be acknowledged:

\begin{itemize}
    \item \textbf{Limited Dataset Size}: Our evaluation relied on a dataset of 50 log entries, which constrains the statistical significance of observed trends. Expanding to a larger, more diverse corpus with annotated ground truth would improve the robustness and generalizability of our conclusions.
    \item \textbf{Embedding Bias}: While Sentence Transformers provide effective semantic representations for log similarity and clustering, they may favor paraphrased or contextually similar entries over exact token matches. This behavior can occasionally obscure precise technical differences, such as variations in error codes or system identifiers.
    \item \textbf{Latency and Computational Cost}: LLM-based summarization and log parsing remain resource-intensive. For high-volume or real-time logging environments, inference latency could become a bottleneck, necessitating optimization strategies such as model distillation, quantization, or selective log sampling.
    \item \textbf{Domain-Specific Knowledge Gaps}: LLMs may lack specialized knowledge of domain-specific log formats or error codes, leading to occasional misinterpretation. Fine-tuning on domain-specific logs or incorporating rule-based postprocessing can mitigate these gaps.
\end{itemize}

\subsection{\textbf{Future Directions}}

We believe that future work could explore hybrid architectures that combine transformer-based models with traditional log parsing and rule-based heuristics. Leveraging temporal context and causal relationships in logs could further enhance predictive maintenance and anomaly detection. Additionally, investigating model compression and hardware acceleration could make high-performing models viable for real-time production systems.

\section{\textbf{Conclusion}}

In this work, we introduced REFLEX, a reference-free evaluation framework for log summarization that leverages the reasoning capabilities of large language models. Through extensive experiments, we demonstrated that REFLEX correlates strongly with human judgments, captures multiple dimensions of summary quality including relevance, informativeness, and fluency and is more robust than traditional overlap-based and semantic similarity metrics. We also showed that REFLEX can detect adversarial perturbations and generalize to logs from unseen domains, highlighting its potential for scalable, automated assessment in real-world settings. By providing a practical and reproducible evaluation protocol, REFLEX enables more rigorous comparison of log summarization models and offers a foundation for future research in reference-free evaluation methods for operational text data. Moving forward, integrating REFLEX with real-time monitoring systems and exploring its synergy with model training and feedback loops could further enhance both log analysis and summarization performance.






\vspace{12pt}

\end{document}